\documentclass[lettersize,journal]{IEEEtran}
\usepackage{amssymb}
\usepackage{amsmath}
\usepackage[english]{babel}
\usepackage{float}
\usepackage{graphicx}
\usepackage[hidelinks]{hyperref}
\usepackage{mathtools}

\usepackage{filecontents}
\usepackage[noadjust]{cite}
\usepackage[font=footnotesize,labelfont=footnotesize]{subcaption}
\usepackage[font=footnotesize,labelfont=footnotesize]{caption}
\usepackage{comment}
\usepackage{authblk}
\usepackage{multirow}
\usepackage{xcolor}
\usepackage{algorithm}
\usepackage{algorithmic}
\usepackage{xspace}
\usepackage{hhline}

\definecolor{darkblue}{rgb}{0.15,0.15,0.55}
\definecolor{lightgrey}{rgb}{0.75,0.75,0.75}

\makeindex             

\begin{document}
\title{Grasping Deformable Objects via Reinforcement Learning with Cross-Modal Attention to Visuo-Tactile Inputs}
\author{Yonghyun Lee$^1$, Sungeun Hong$^2$, Min-gu Kim$^3$, Gyeonghwan Kim$^1$, and Changjoo Nam$^{1,*}$
\thanks{This work was supported by the National Research Foundation of Korea (NRF) grant funded by the Korea government (MSIT)(No. RS-2024-00461583). $^1$Dept. of Electronic Engineering at Sogang University, Seoul, Korea. $^2$Dept. of Immersive Media and Engineering at Sungkyunkwan University, Seoul, Korea. $^3$College of Medicine, Yonsei University, Seoul, Korea. $^*$\textit{Corresponding author: Changjoo Nam} ({\tt\small cjnam@sogang.ac.kr})}%
}


\maketitle
\thispagestyle{empty}
\begin{abstract}
We consider the problem of grasping deformable objects with soft shells using a robotic gripper. Such objects have a center-of-mass that changes dynamically and are fragile so prone to burst. Thus, it is difficult for robots to generate appropriate control inputs not to drop or break the object while performing manipulation tasks. Multi-modal sensing data could help understand the grasping state through global information (e.g., shapes, pose) from visual data and local information around the contact (e.g., pressure) from tactile data. Although they have complementary information that can be beneficial to use together, fusing them is difficult owing to their different properties. 

We propose a method based on deep reinforcement learning (DRL) that generates control inputs of a simple gripper from visuo-tactile sensing information. Our method employs a cross-modal attention module in the encoder network and trains it in a self-supervised manner using the loss function of the RL agent. With the multi-modal fusion, the proposed method can learn the representation for the DRL agent from the visuo-tactile sensory data. The experimental result shows that cross-modal attention is effective to outperform other early and late data fusion methods across different environments including unseen robot motions and objects.
\end{abstract}


\section{Introduction}

\IEEEPARstart{H}{uman} beings are extremely skillful at manipulating objects owing to their ability to agglomerate multi-modal sensory inputs together for controlling their hands. As robotic technologies advance, robots also can manipulate various objects using their sensory inputs~\cite{palleschi2023grasp}. While previous research has focused on manipulating rigid objects~\cite{bekiroglu_probabilistic_2013, hebert_fusion_2011, joshi_robotic_2020, calandra_more_2018}, methods for handling deformable objects have received much attention recently~\cite{cui_grasp_2020, han_learning_2023, pecyna2022visual}. However, manipulating soft deformable objects is still challenging to robots because such objects have center-of-masses and shapes that change dynamically. Although dexterous hands are capable of handling deformable objects with precision, their high costs have limited their widespread use. Therefore, it is essential to develop a method that can create robust control inputs using different sensing modalities for simpler robots to manipulate deformable objects.

\begin{figure}[t!]
    \centering
    \includegraphics[width=0.33\textwidth]{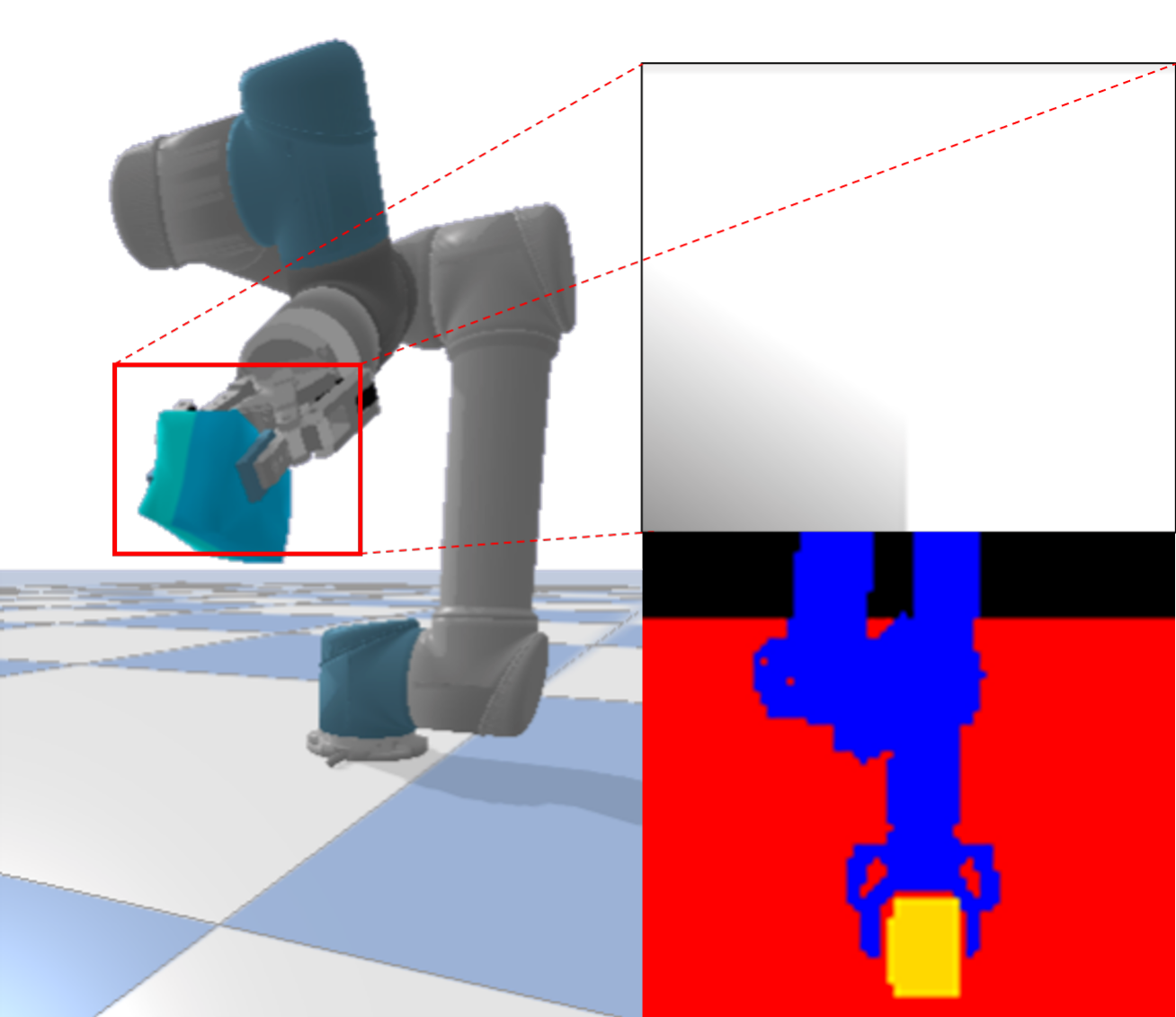}
    \caption{The grasping task of deformable objects. (Left) A two-jaw robotic gripper attached to a 6-DOF arm. While the arm moves, the goal is to generate the control input to the gripper to grasp the object stably and safely without dropping or breaking it. (Upper right) A concatenated image of two tactile images obtained from the sensors attached to the inside of the finger tips. The tactile data provide local information around the contact area.  (Lower right) A segmentation mask from an RGB image. The visual data can provide global information about grasping. }
    \label{fig:example}
\end{figure}

Among various manipulation tasks, we consider grasping which is one of the primary tasks to work with soft deformable objects as shown in Fig.~\ref{fig:example}. There have been a line of research on predicting the current grasp states based on supervised learning approaches using visual and/or tactile information~\cite{han_learning_2023,bekiroglu_probabilistic_2013,calandra_more_2018} in order to use the states for grasping. However, predicting accurate grasp states through supervised learning requires a significant amount of data. Also, they need an additional method to generate control inputs of robot hands based on the prediction result. On the other hand, reinforcement learning (RL) frameworks are able to generate control inputs to robots in an end-to-end fashion from sensor data.

Deep reinforcement learning (DRL) frameworks have brought significant progresses to robotic manipulation where continuous action spaces are prevalent. Among these frameworks, policy gradient approaches are widely used. An actor network predicts actions based on the given information of the environment while a critic network evaluates the sampled actions to facilitate learning~\cite{konda1999actor}. As directly using high-dimensional visual and tactile data is inefficient, the use of an encoder becomes crucial in order to transform these sensor data into compact features that encapsulate essential information. 

Few DRL algorithms have been proposed to encode high-dimensional visual and tactile information for efficient learning~\cite{hansen_visuotactile-rl_2022,pecyna2022visual,calandra_more_2018,cui_grasp_2020,han_learning_2023}. However, they either separately use the two different sensing modalities for a sequential decision-making~\cite{hansen_visuotactile-rl_2022} or simply concatenate feature vectors obtained separately from the each encoder networks composed of several convolution layers~\cite{pecyna2022visual, calandra_more_2018, cui_grasp_2020, han_learning_2023}. Such simple modality fusion methods may not be successful in learning more meaningful representations as they could not fully exploit the complementary information from different sensing modalities.

We propose a DRL framework that combines an encoder with a cross-modal attention module to fuse the complementary information effectively from tactile and visual sensors (the right images in Fig.~\ref{fig:example}). Cross-modal attention has been applied to RGB-D and RGB-T crowd counting~\cite{zhang_spatio-channel_2022} successfully to selectively focus on relevant features from different modality. However, the different modality data used in~\cite{zhang_spatio-channel_2022} capture the same spatial area by the RGB, depth, and thermal cameras with the same field of view. On the other hand, visual and tactile information have different sensing areas which are complimentary. Also, our DRL agent learns to act from the interactions with the object so has no explicit label to train the encoder unlike the supervised learning problem. 

To the best of our knowledge, it is the first attempt to use an encoder network with a cross-attention module for a DRL agent to learn the representation from visuo-tactile information. We employ a cross-modal attention module proposed in~\cite{zhang_spatio-channel_2022} which enables the RL agent to selectively focus on relevant features from each modality to better understand the correlation between tactile and visual inputs. For better exploration, we employ a Soft Actor-Critic (SAC) algorithm~\cite{haarnoja2018soft} with random data augmentation~\cite{yarats2021mastering}. The training environment of the agent is constructed using PyBullet~\cite{coumans2021}, which is a dynamic simulator. In the environment, we implement various shapes of deformable objects and tactile sensors using TACTO~\cite{Wang2022TACTO}. Our experimental result shows that the proposed method outperforms other sensor fusion methods (i.e., early fusion, late fusion) across different environments including unseen robot motions and objects. 

Our main contributions are as follow:
\begin{itemize}
    \item We develop an end-to-end DRL framework that can fuse visuo-tactile information effectively through a cross-attention mechanism. Our method is able to generate continuous control inputs to a simple robotic gripper to grasp soft deformable objects safely while not dropping or breaking them. 
    \item We develop a training and test suite using a dynamic simulator. The suite can import different robot models whose gripper is attached with tactile sensors. It can simulate deformable objects with varying shapes and properties such as mass, modulus, and friction.
    \item We provide experimental results comparing RL methods that use multi-modal sensing data without learning representations. Our method shows better performance in terms of the cumulative return and the actual grasping task in seen and unseen environments. 
\end{itemize}

\section{Related Work} 
As deep learning technologies enabled handling of high-dimensional data, there has been much attention to grasping deformable objects in an end-to-end fashion using raw sensor data such as RGB images and tactile values. Unlike the methods that evaluate grasps to find a suitable grasp configuration~\cite{calandra_more_2018,feng2020center,han_learning_2023}, directly controlling a robot end-effector can reduce errors from inaccurately assessing grasp states albeit difficult.

Existing works mostly use vision-only~\cite{jangir_dynamic_2020,matas_sim--real_2018,scheikl_sim--real_2023,seita2021learning}, tactile-only~\cite{kaboli2016tactile}, or vision and tactile together~\cite{hansen_visuotactile-rl_2022,pecyna2022visual} to control the end-effector directly. The two sensing modalities provide complementary information about grasping. While visual information captures the global state of objects including pose and deformation, tactile information provides dense local information at contact. Force-torque sensors also can be useful to recognize the movement of deformable objects~\cite{lee_making_2019} but they cannot capture the deformation of objects and expensive (several thousands US dollars).

We focus on the methods which use visual and tactile information together~\cite{hansen_visuotactile-rl_2022,pecyna2022visual} to leverage the complementary modalities for safe and reliable grasping. In~\cite{han_learning_2023}, the authors propose a grasping framework that uses Transformer models to predict a grasp force given sequential visual and tactile information of an object. Although the framework can process spatial-temporal features and capture long-term dependencies in sequential data, the inference time (2--3 Hz) may not be satisfactory for robotic tasks. Also, the Transformer models do not directly capture the interactions between the robot and the object since the sequential actions applied to the object are not explicitly incorporated. 

In~\cite{hansen_visuotactile-rl_2022}, a DRL algorithm is proposed to use pixel-level visual and tactile feedback. Since RL represents an environment using a Markov Decision Process (MDP), the interactions between the robot and object and their outcomes are explicitly described by the transition and reward functions. The algorithm implements tactile gating, tactile data augmentation, and visual degradation to deal with changes in lighting conditions, camera views, weights, and friction of objects. Although using the two sensing modalities enables the algorithm to outperform visual-only and tactile-only methods, the visual and tactile data are simply combined without selective fusion. Specifically, the visual and tactile images are combined on the channel axis to be used by a single convolutional encoder. In another setting, the algorithm concatenates the feature maps from two separate encoding networks for each of the visual and tactile images. Since the data from two different sensing modalities have different sensing regions, properties, and implications, simply combining them may miss the chance of learning the structure of the data. Moreover, the algorithm is not tested with deformable objects but only rigid ones. 

Another DRL algorithm is proposed to use visual and tactile information to manipulate deformable linear objects (e.g., rope)~\cite{pecyna2022visual}. The raw tactile values are processed to estimate the geometry (i.e., angle and position) of the rope in contact. The RGB image is also used to extract the angle and position of the rope. They consist of the observations along with the proprioceptive information of the gripper. While these extracted features can provide useful information to control the gripper, such extraction processes usually involve estimation errors. Also, the algorithm does not selectively fuse the two sensing modalities like~\cite{hansen_visuotactile-rl_2022}. 

In order to learn the structure of the multi-modal sensing data, we use the encoder with a Cross-modal Spatio-Channel Attention (CSCA) module proposed in~\cite{zhang_spatio-channel_2022} in order to selectively focus on relevant features from each sensing modality. We use the learned representation as the input information for a DRL algorithm. The combination of such a data fusion mechanism and an RL agent is the first attempt to the best of our knowledge.

\section{Background} 

\subsection{Markov Decision Processes}
To formulate our image-based multi-modal control problem, we use an MDP defined as a tuple $(\mathcal{S},\mathcal{A},p,r,\gamma)$ where $\mathcal{S}$ and $\mathcal{A}$ are the sets of continuous states and actions, respectively. The transition dynamics of the environment $p = p(\textbf{s}_{t+1}|\textbf{s}_{t},\textbf{a}_{t})$ capture the probability distribution given the current state $\textbf{s}_{t} \in \mathcal{S}$ and action $\textbf{a}_{t} \in \mathcal{A}$. The reward function $r:\mathcal{S}\times \mathcal{A} \to \mathbb{R}$ is determined by the current state and action. The quantity $\gamma \in [0,1)$ is a discount factor to emphasize rewards in a near feature rather than rewards in the distant future. A sequence of states $\mathcal{T}$ is called a trajectory or an episode and stored in the experience replay buffer~\cite{mnih2013playing}. The objective is to learn a policy $\pi(\textbf{a}_{t}|\textbf{s}_{t})$ that maximizes the cumulative discounted return $\mathbb{E}_{\pi}[\sum_{t=1}^{\infty} {\gamma^t r_t|\textbf{a}_{t} \sim \pi(\cdot | \textbf{s}_{t}), \textbf{s}_{t+1}\sim p(\cdot | \textbf{s}_{t}, \textbf{a}_{t}), \textbf{s}_{1}\sim p(\cdot)}]$.

\subsection{Algorithm}
We adopt data augmentation techniques used in Data Regularized Q (DrQv2)~\cite{yarats2021mastering} for visual DRL. Although DrQv2 implements Deep Deterministic Policy Gradient (DDPG)~\cite{lillicrap2015continuous}, we replace DDPG by SAC which learns a stochastic policy by maximizing the expected returns and entropy of the policy leading to more exploration~\cite{ziebart2008maximum}. 
Through SAC, we parameterize the state-action value function $Q_\theta(\textbf{s}_{t},\textbf{a}_{t})$, a stochastic policy network $\pi_{\phi}(\textbf{a}_{t}|\textbf{s}_{t})$, and a temperature $\alpha$ to find an optimal policy. The state-action value function parameters $\theta$ can be trained to minimize the soft Bellman residual:
\begin{equation}
\label{eq:JQ1}
J_Q(\theta) = \mathbb{E}_{(\textbf{s}_{t},\textbf{a}_{t}) \sim \mathcal{T}}[\frac{1}{2}(Q_\theta (\textbf{s}_{t}, \textbf{a}_{t}) - y_t)^2]
\end{equation}
where
\begin{eqnarray}
\label{eq:JQ2}\nonumber
y_t = r(\textbf{s}_{t},\textbf{a}_{t}) + \gamma(\textbf{s}_{t+1},\textbf{a}_{t+1})  \\ 
 - \alpha \log (\pi_{\phi}(\textbf{a}_{t+1}|\textbf{s}_{t+1})).
\end{eqnarray}
The policy network parameters $\phi$ can be learned by minimizing the objective:
\begin{equation}
\label{eq:JPI}
J_\pi(\phi) = \mathbb{E}_{\textbf{s}_{t} \sim \mathcal{T}, \textbf{a}_{t} \sim \pi_\phi}[\alpha \log (\pi_\phi(\textbf{a}_{t}|\textbf{s}_{t})) - Q_\theta (\textbf{s}_{t}, \textbf{a}_{t}) ]
\end{equation}
When the training is going on, (\ref{eq:JQ1}), (\ref{eq:JPI}) are minimized with gradient descent.

\subsection{Encoder network}
\label{sec:encoder}
\begin{figure*}
    \centering
    \includegraphics[width=0.88\textwidth]{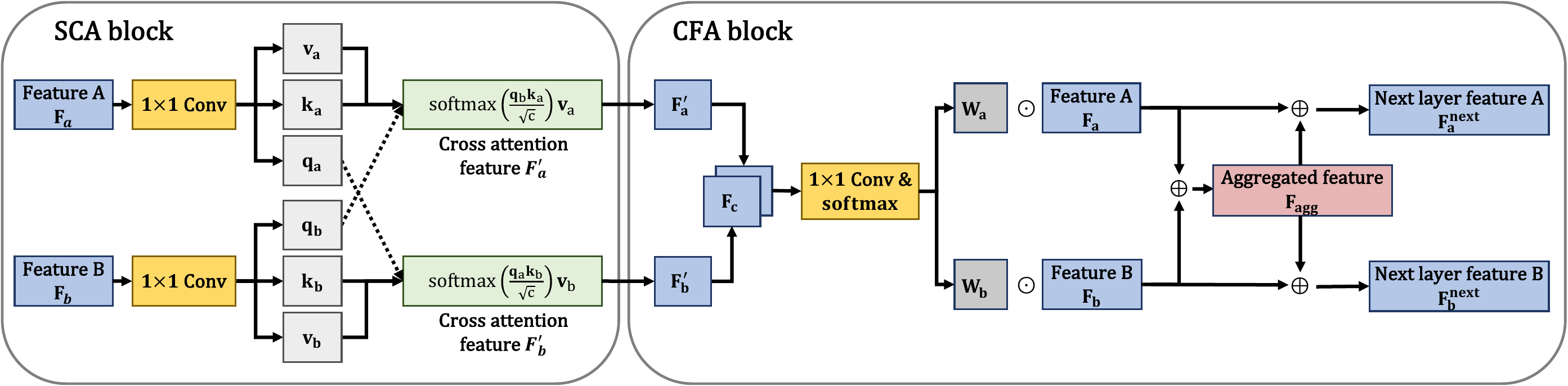}
    \caption{The architecture of the cross-modal attention module proposed in~\cite{zhang_spatio-channel_2022}. The module consists of SCA block to extract global feature correlations of the multi-modal data, and the CFA block to integrate complementary features.}
    \label{fig:csca}
\end{figure*}

We use Cross-modal Spatio-Channel Attention (CSCA) module~\cite{zhang_spatio-channel_2022} to selectively focus on the complementary information from multi-modal sensor data. Through the module, visual and tactile inputs are fused to embed the aggregated features and then converted to the same dimensional outputs using skip connection. The module consists of two operations: Spatial-wise Cross-modal Attention (SCA) block and Channel-wise Feature Aggregation (CFA) block as described in Fig.~\ref{fig:csca}.

The SCA module is designed to capture global feature correlations of the multi-modal inputs based on the cross-modal attention mechanism (the left half of Fig.~\ref{fig:csca}). First, each of the multi-modal inputs is projected separately into three vectors: Query $\textbf{q}$, Key $\textbf{k}$, and Value $\textbf{v}$ through a $1 \times 1$ convolution. Unlike non-local neural network~\cite{wang2018non} and self-attention mechanism~\cite{vaswani2017attention}, the cross-modal attention mechanism involves the query, key, and value, where the key and value come from one modality while the query originates from the other modality. During the process of optimizing the CSCA module, the SCA block integrates the information from two modalities.

From the complementary information of the multi-modal data, the attention module generates an aggregated feature in the CFA block (the right half of Fig.~\ref{fig:csca}). The cross-modal attention results $\textbf{F}_{a}^\prime$ and $\textbf{F}_{b}^\prime$ are concatenated into a feature map $\textbf{F}_{c}$. A $1 \times 1$ convolution and a softmax operation follow to produce weight maps $\textbf{W}_{a}$ and $\textbf{W}_{b}$ to re-weight the original feature maps:
\begin{equation}
[\textbf{W}_{a},\textbf{W}_{b}] = \text{softmax}(\text{MLP}(\textbf{F}_{c})).
\end{equation}

Each of the weight maps are multiplied element-wise with the original feature maps. The weighted feature maps are summed to generate the final aggregation feature map $\textbf{F}_{agg}$ that is
\begin{equation}
\textbf{F}_{agg} = \textbf{W}_{a} \odot \textbf{F}_{a} + \textbf{W}_{b} \odot \textbf{F}_{b}.
\end{equation}

If the CSCA module is used between each layer of a CNN, the updated feature maps are passed to the next layer of the CNN according to
\begin{equation}
\textbf{F}_{a}^{next} = (\textbf{F}_{agg} + \textbf{F}_{a})/2,\qquad \textbf{F}_{b}^{next} = (\textbf{F}_{agg} + \textbf{F}_{b})/2.
\end{equation}
If no more layer of CNN follows, the final aggregation feature map is used in the subsequent steps as a representation fusing selective information from each modality.


\section{Method}
In this section, we describe our proposed design of the DRL agent and the network architecture. The agent learns the state representation through the encoder network described in Sec.~\ref{sec:encoder}. Since RL does not have a fixed dataset with labels, we cannot train the encoder in a supervised fashion. Thus, we train it using the backpropagation that computes the gradient of the objective function (\ref{eq:JQ1}). Our training environment is built using a dynamic simulator PyBullet~\cite{coumans2021} to facilitate convenient and efficient training.

\subsection{Agent description}

We focus on a grasping task where a simple two-jaw robotic gripper maintains a stable grasp of a soft-shell deformable object while a high-DOF robotic arm connected to the gripper moves. The gripper has one control input (velocity or position) to grasp the object. In our MDP, action $\textbf{a}_{t}= \textbf{v}_{t} \in \mathcal{A}$ represents the velocity of the joints of the gripper at time step $t$. A state $\textbf{s}_t \in \mathcal{S}$ is defined to include the aggregated feature from the encoder network and the proprioceptive information of the robot. Specifically, $\textbf{s}_t = [\textbf{s}_f, \textbf{s}_r]$ where $\textbf{s}_f$ is the embedding vector from the encoder network. The proprioceptive information is included in $\textbf{s}_r$ such as the 6-D pose of the end-effector, the joint position, and the velocity of the gripper. The dimensions of the state vectors are determined depending on the specific encoder network and the robot used.

\begin{figure}
    \centering
    \includegraphics[width=0.28\textwidth]{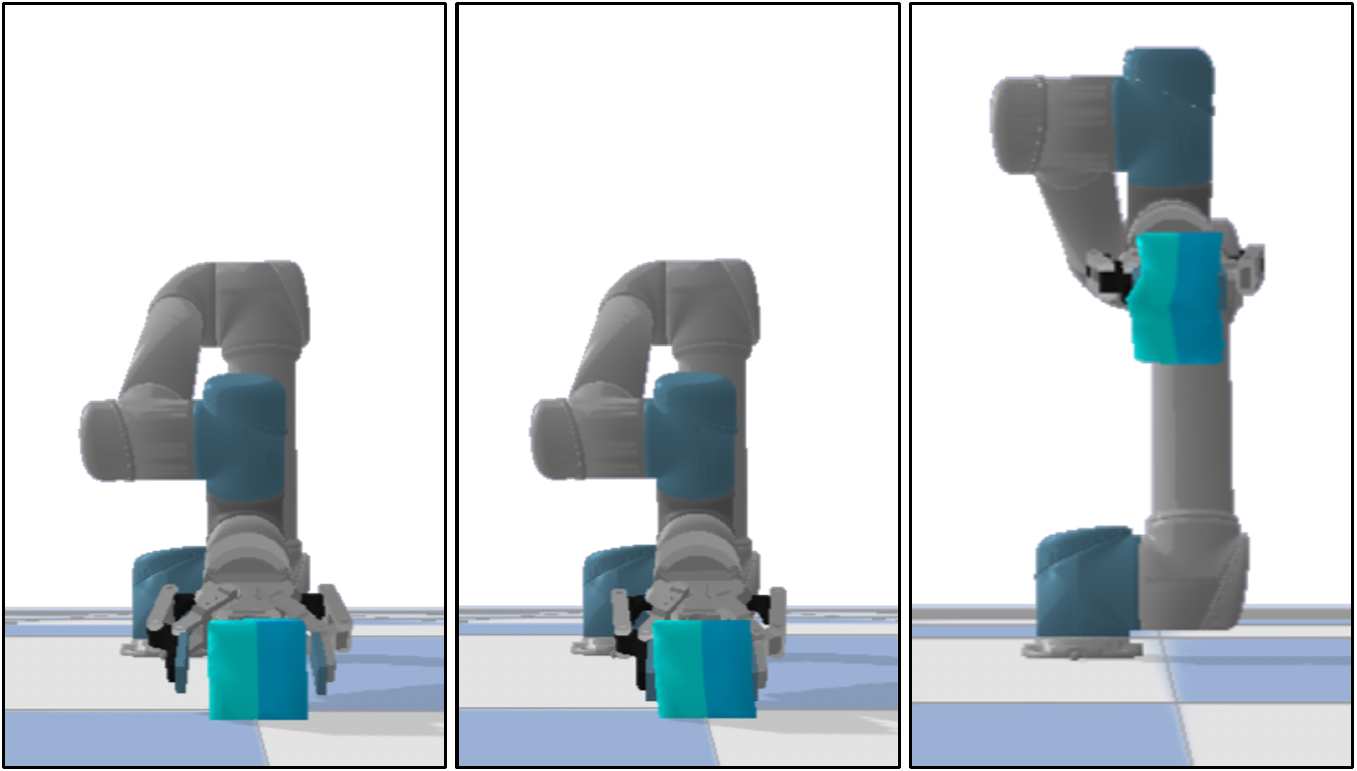}
    \caption{An example of the grasping task. (Left) The arm is controlled to move the gripper down to the object. (Mid) The agent begins to control the gripper to grasp the object. (Right) The arm is controlled to lift its gripper.}
    \label{fig:Taskorder}
\end{figure}

An episode starts with the gripper located above a deformable object on a floor. An example task is described in Fig.~\ref{fig:Taskorder}. In the initial stage, the arm is controlled to move the gripper down to the object. Once the gripper is aligned with the object, the agent begins to control the gripper to grasp the object on the floor. After a few steps, the arm is controlled to lift its gripper without any feedback from the current state of the object. If the gripper holds the object until it arrives at the goal location (up in the 30cm above the start location), the episode terminates successfully. An episode fails if the gripper loses hold of the object or applies excessive force which may lead to irreversible damage to the deformable object. 

The reward function is designed to encourage the agent to apply an appropriate control input to the gripper while the arm moves: 
\begin{equation} \label{eq:reward}
    r(\textbf{s}_{t}, \textbf{a}_{t}) = r_{s}^{t} + r_{f}^{t} + r_{c}^{t}
\end{equation}
where $r_{s}^{t}$ and $r_{f}^{t}$ are the rewards received when an episode terminates successfully and fails, respectively. In addition to the sparse rewards, a continuous reward $r_{c}^{t}$ is given to provide more guidance to the agent during training. Specifically, 
\begin{equation} \label{eq:reward_s}
    r_{s}^{t} = \begin{cases*}
        1   & if the object reaches the goal,\\
        0          & otherwise
    \end{cases*}
\end{equation}
and 
\begin{equation} \label{eq:reward_f}
    r_{f}^{t} = \begin{cases*}
        -0.3  & if the object is broken,\\
        -0.7  & if the object falls,\\
        0          & otherwise
    \end{cases*} 
\end{equation}
where the penalty for dropping the object is more harmful to learning than the penalty for breaking the object. Because we consider that closing the gripper is more important than spreading to reach the goal with stable grasping the deformable object. The continuous reward is defined as
\begin{equation} \label{eq:reward_c}
    r_{c}^{t} = \begin{cases*}
        P/T  & if an episode is not terminated,\\
        0          & otherwise
    \end{cases*} 
\end{equation}
where $P$ is the portion of tactile readings that are at least 30\% of the maximum tactile value. Given a tactile image whose size is $H \times W$, 
$$P = \frac{\sum_{i,j} x_{ij}}{H \times W}$$
for $i = 1, \cdots, W$ and $j = 1, \cdots, H$ where $x_{ij} = 1$ if the corresponding tactile value (with the indices $i$ and $j$) is greater than or equal to $0.3$ (notice that the tactile values are normalized).
On the other hand, $T$ is the maximum episode step referring to the total number of steps it takes for the robotic arm to reach the goal location from the initial stage. 
The proportion of the tactile sensor value encourages the gripper to grasp the object more stably by guiding the fingers to grip with more contact areas. Ideally, the total sum of rewards that can be maximally obtained through continuous rewards is 1. Then the maximum expected return $\sum_{t}^{T}{r(\textbf{s}_{t}, \textbf{a}_{t})}$ is 2. 

\subsection{Model description}

\begin{figure*}
    \centering
    \includegraphics[width=\textwidth]{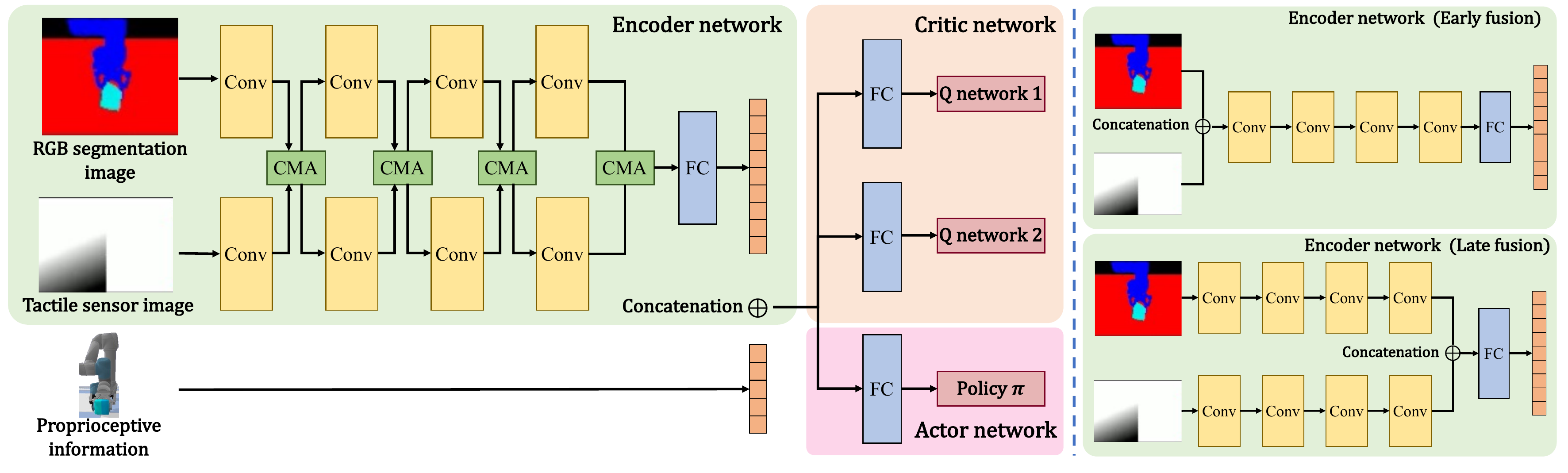}
    \caption{(Left) The overall structure of the cross-attention DRL framework (DRQ-CMA). The encoder learns the representation for the RL agent from the visuo-tactile information by selectively focusing on relevant data. (Right) The encoders used in DRQ-EF and DRQ-LF without the attention mechanism.}
    \label{fig:model}
\end{figure*}

The overall structure of our multi-modal RL agent is shown in Fig.~\ref{fig:model}. The architecture consists of three components: the encoder network, a critic network, and an actor network. The encoder network maps an RGB image and a one-channel tactile image into separate feature maps. 

For the visual data, we use semantic segmentation masks with the semantic labels of the object, robot, and backgrounds. Compared to using raw images, segmentation masks enable the agent to focus on the shape of the object and the move of the robot without dependency on visual patterns. As a result, changes in the object, robot model, or background could cause minimal impacts on the performance of the agent as long as they can be semantically labeled.\footnote{Foundation models such as Segment Anything~\cite{kirillov2023segment} can accelerate the learning to segment new objects.}

Through the intermediate and final CSCA modules, the encoder generates an aggregated feature. The feature is then transformed into a $1 \times 50$ visual embedding $\textbf{s}_f$, through a fully connected layer. The embedding vector is concatenated with a $1 \times 7$ vector $\textbf{s}_r$ including the proprioceptive data of the robot. The final $1 \times 57$ transformed embedding vector $\textbf{s}_t$ is the input to the critic and actor networks.

The critic network estimates the expected return of the current state. While training the critic network, the encoder network is also trained through backpropagation using the objective function (\ref{eq:JQ1}). As a result, the encoder can be capable of distinguishing each state based on the expected return of the critic network and the actor network can find the optimal action policy for that state. By leveraging the CSCA module, the encoder network helps the critic and actor networks understand better the current state which is the learned representation of the high-dimensional multi-modal sensor data.

\section{Experiments}  

In this section, we show the training and test results of the methods with different fusion schemes. They are (i) DRQ with an early fusion encoder (DRQ-EF), (ii) DRQ with a late fusion encoder (DRQ-LF), (iii) our proposed method, DRQ with the cross-modal attention module encoder (DRQ-CMA). In order to evaluate the effect of two sensing modalities, we also test a method using only the visual information, which is DRQ with an encoder for segmentation image only (DRQ-SO). 

In DRQ-EF, the tactile and segmentation images are simply concatenated on the channel axis. In DRQ-LF, the images are separately passed through the convolutional layer of their own encoder network. The two resulting feature maps are then concatenated and passed through a fully connected layer to produce a final $1 \times 50$ embedding vector. The two different encoder networks are briefly described in the left of Fig.~\ref{fig:model}.

\begin{figure}
    \centering
    \includegraphics[width=0.35\textwidth]{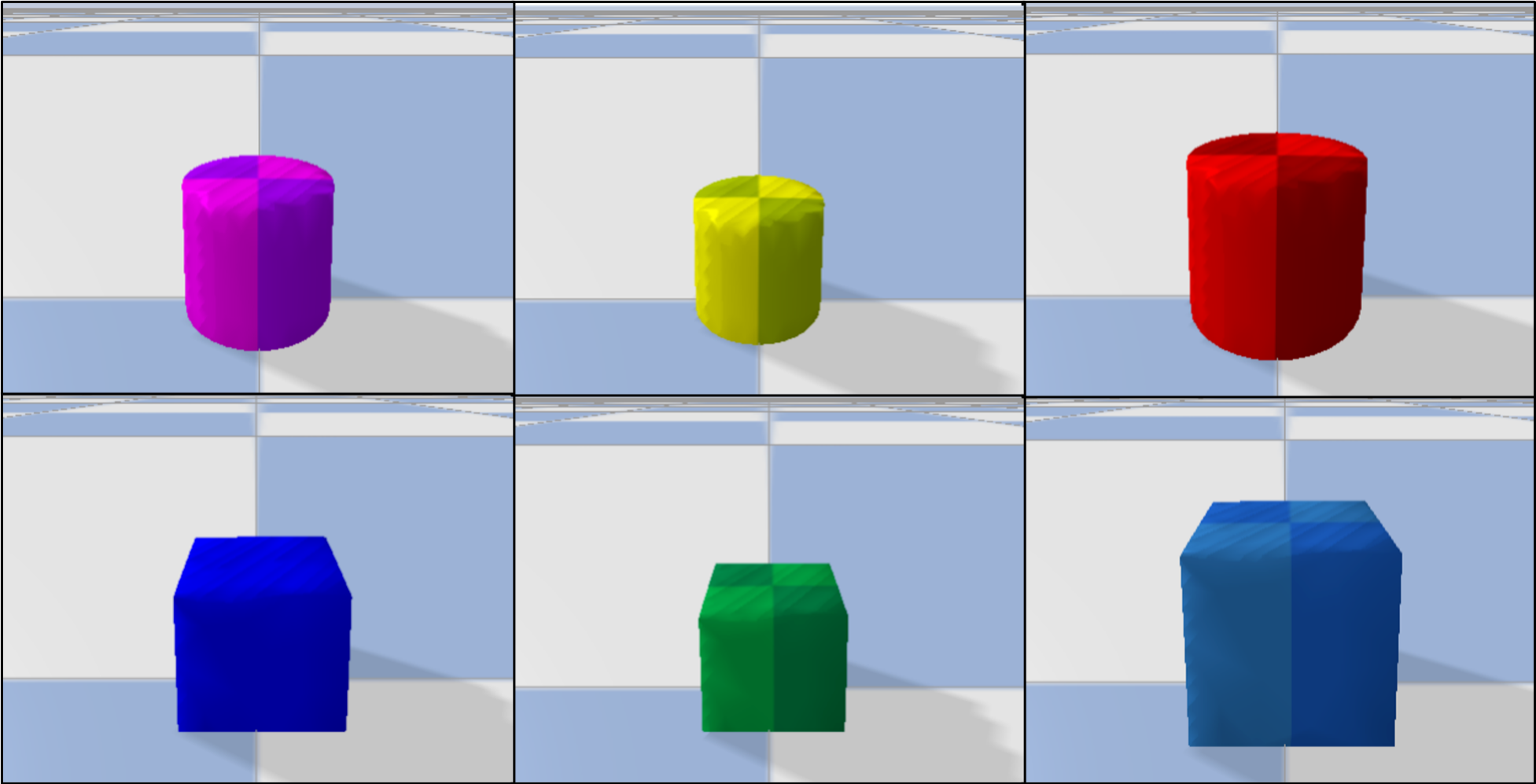}
    \caption{The deformable objects used in training and test. The two left objects are used in training. In tests, all objects are used including the four unseen objects. (Left) A cylindrical object and a hexahedron which weigh 0.5 kg. (Mid) The sizes of the objects are 0.8 times of the left ones. Weights are 0.25 kg for both. (Right) The sizes are 1.2 times of the left ones. Their weights are 2 kg.}
    \label{fig:objects}
\end{figure}

The PyBullet simulation environment is used in both training and test. In the simulation, we import a 6-DOF robotic arm equipped with a Robotiq 2F-85 gripper. We use six soft deformable objects with two shapes, three sizes, three weights, and colors as shown in Fig.~\ref{fig:objects}. Two of them (in the left of Fig.~\ref{fig:objects}) are used in training of all methods with different fusion schemes. They are also used in tests along with the rest four unseen objects in Fig.~\ref{fig:objects} to show the generalization performance of the compared methods. We run experiments using tactile and segmentation images with a dimension of $128 \times 128$ same as the RGB image used in DRQ~\cite{yarats2021mastering}. Note that the dimensions of the images can change depending on the output from the tactile sensors and segmentation network that are employed.

We use two metrics to evaluate the performance of the methods. They are the batch reward during the training process and the success rate of the grasping task in the evaluation stage. The batch reward measures the performance during the training process. It is the average reward obtained from the $256$ MDP tuples sampled from the replay buffer at each step during training. The success rate is used to evaluate the performance of the learned policy with the grasping task. It measures how many successful test episodes are achieved out of 20 trials. In a successful episode, the object is lifted to the goal location (30 cm above the floor) without dropping or breaking the object. 

Across all methods, the number of training episode steps is $450$ million. The AdamW optimizer is used with a learning rate of $10^{-4}$. The batch size is $216$. A random exploration phase is performed in the first $20,000$ steps. All experiments are done in a system with Intel Core i9-10920X CPU 3.50 GHz, NVIDIA GeForce RTX 3090Ti GPU, and 256GB RAM.

\begin{figure}
    \centering
    \includegraphics[width=0.45\textwidth]{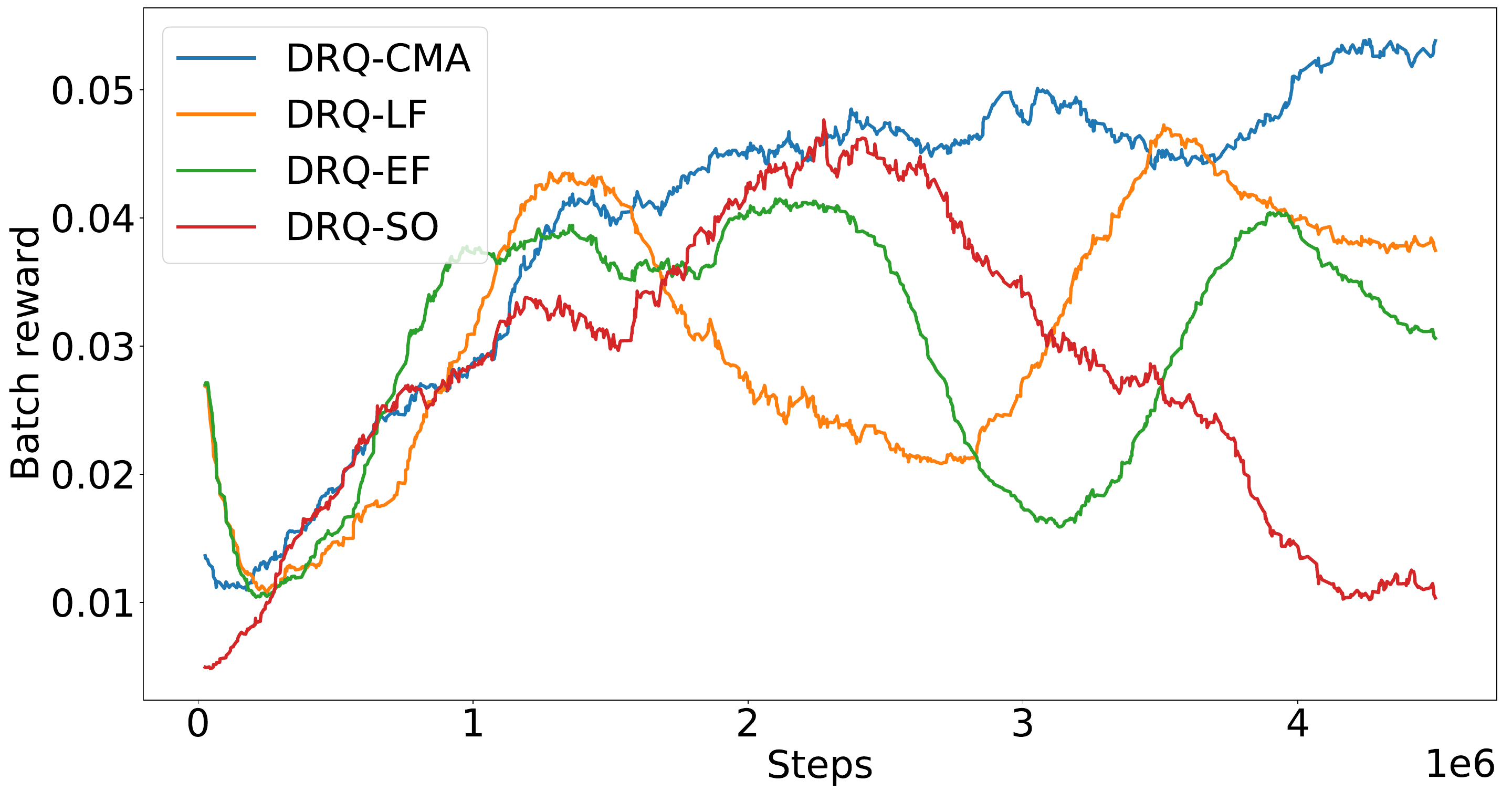}
    \caption{The batch reward obtained during training. Only DRQ-CMA shows an upward trend, which demonstrates the ability to learn the representation from the cross-modal attention.}
    \label{fig:BR}
\end{figure}

\subsection{Training process}

As the agent experiences more episodes with successful attempts to complete the task, the batch reward steadily increases as shown in Fig.~\ref{fig:BR}. The batch reward of DRQ-CMA shows an upward trend which indicates that the learning process is steadily optimized. However, the learning process of DRQ-EF and DRQ-LF are less stable as the batch reward drops and fluctuates after the initial rise. DRQ-SO is not able to learn as the batch reward continues to decrease as training proceeds. 

The comparison between DRQ-SO and ours (DRQ-CMA) clearly shows the benefit of using visual and tactile data both. Although visual information can provide the shapes of the object being grasped, the agent is not aware of the dynamics at the contact area between the robot and the object. Tactile information has been shown to be an appropriate complementary modality for the grasping task of deformable objects.

On the other hand, the comparison between our method and other fusion methods shows the effectiveness of the cross-modal attention. Although the batch reward of DRQ-EF and DRQ-LF resumes increasing after the initial drops, they are not able to have continuously increasing performance. Nevertheless, their performance outperforms DRQ-SO using a single sensing modality.

\subsection{Evaluation results}


We evaluate the learned policy of the four methods with grasping tasks. To see the generalization performance of the policies, we set three different test environments, which have different move patterns of the robot arm (\textit{random move}) or different sets of deformable objects (\textit{more objects}). They are compared with a baseline environment (\textit{basic}) which has the identical setup to the training environment. Each method in each environment is tested over 20 trials to calculate the success rate. 

In \textit{basic} environment, the test episodes consist of grasping the two objects shown in the left of Fig.~\ref{fig:objects}. Also, the robot gripper moves vertically by following a straight path to the goal location. In \textit{random move} environment, the set of test objects is the same as \textit{basic}. The robot arm moves differently by randomly choosing three way points until the end-effector arrives at a random goal location. As a result, the end-effector grasping the object shows zigzag moving patterns. In \textit{more objects} environment, other settings are the same with \textit{basic} except for the test objects including all objects shown in Fig.~\ref{fig:objects}. They are randomly chosen in the 20 test trials. 

\begin{figure}
    \centering
    \includegraphics[width=0.5\textwidth]{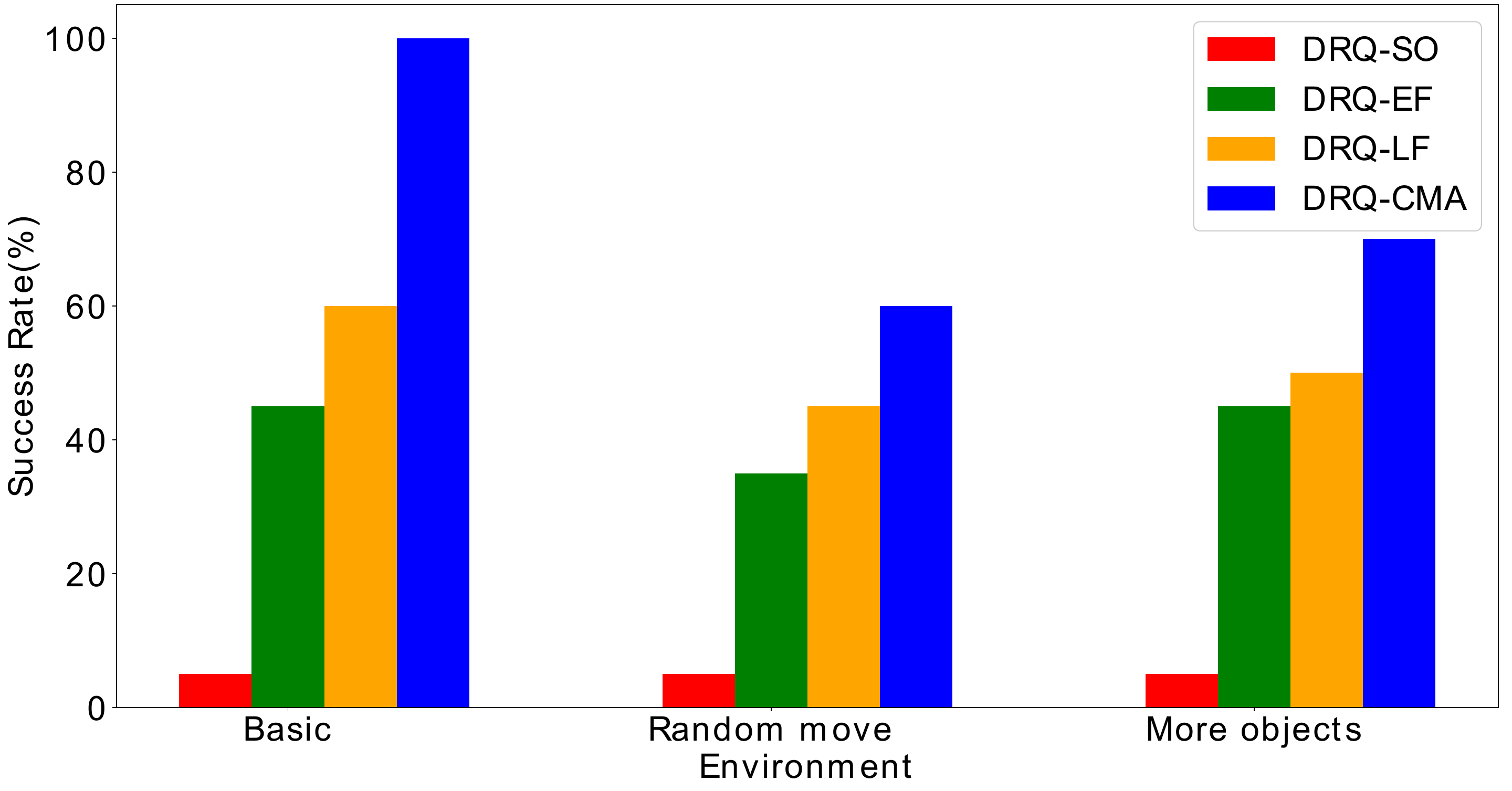}
    \caption{The success rate of the three environments. First, the basic environment is same environment during training process. Secondly, the random move environment has various trajectories with an arbitrary intermediate point. Lastly, the more objects environment has various objects appearing one by one.}
    \label{fig:SR}
\end{figure}

The result is summarized in Fig.~\ref{fig:SR}. DRQ-SO mostly fails at the grasping task across different environments, which is expected from the training result. Even in the same environment where training is done, the success rate is low (5\%). Random moves of the arm seem to be more difficult than handling different objects. It is harder to maintain a stable grasp if the dynamics of the object fluctuate. Grasping unseen objects is better handled probably because of tactile values which can provide meaningful information about the pressure in the contact regardless of the shapes. 

The success rate of DRQ-CMA drops up to 40\% while others have lesser drops. The large performance drop of DRQ-CMA could be caused by overfitting. DRQ-CMA has more learnable parameters, both the encoder network and the trained policy could be overfitted to the trajectories occurred in the training environment. Others are with less learnable parameters so their performance drops can be moderate. Nonetheless, the success rate of DRQ-CMA is the highest across all environments. Also, it shows a certain level of generalization ability even though the training episodes are seldom randomized.


\section{Conclusion} 
We proposed an end-to-end multi-modal DRL framework that fuses visual and tactile information to grasp deformable objects. The proposed method can learn the policy generating the control input to a two-jaw simple gripper as well as the representation for the RL agent from multi-modal sensory data. From experiments, we showed that the cross-modal attention mechanism for data fusion is effective to provide an appropriate representation for the RL agent. Our method outperformed the methods with a single sensing modality and those with simple fusion schemes across different environments including unseen robot motions and objects. 

We identified several future directions. We will work on the overfitting problem of the proposed method by applying various techniques such as regularization, dropout, and curriculum training with various objects and motions of the robot. Domain randomization is also a key for generalization and the sim-to-real transfer. We will also consider more experiments to show the generalization performance. Lastly, we are in progress to build a physical robot system with tactile sensors to apply the proposed method. 


\bibliographystyle{IEEEtran}
\bibliography{references}
\end{document}